\documentclass[final]{IEEEtran}

\usepackage[noadjust]{cite}

\usepackage{graphicx}

\usepackage{subcaption}

\usepackage{amssymb}

\usepackage{color}

\usepackage[normalem]{ulem}

\usepackage{amsmath}
\usepackage{bm}

\usepackage[group-separator={,}]{siunitx}

\begin{document}

\title{Whole-Slide Mitosis Detection in H\&E Breast Histology Using PHH3 as a Reference to Train Distilled Stain-Invariant Convolutional Networks}

\author{David~Tellez*, Maschenka~Balkenhol, Irene~Otte-H\"oller, Rob~van~de~Loo,
		Rob~Vogels, Peter~Bult, Carla~Wauters, Willem~Vreuls, Suzanne~Mol, Nico~Karssemeijer,
        Geert~Litjens, Jeroen~van~der~Laak, Francesco~Ciompi
\thanks{*D. Tellez is with the Diagnostic Image Analysis Group and the Department of Pathology, Radboud University Medical Center, 6500HB Nijmegen, The Netherlands (e-mail: david.tellezmartin@radboudumc.nl).}
\thanks{M. Balkenhol, I. Otte-H\"oller, R. van de Loo, G. Litjens, J. van der Laak, and F. Ciompi are with the Diagnostic Image Analysis Group and the Department of Pathology, Radboud University Medical Center, 6500HB Nijmegen, The Netherlands.}
\thanks{R. Vogels, and P. Bult are with the Department of Pathology, Radboud University Medical Center, 6500HB Nijmegen, The Netherlands.}
\thanks{C. Wauters, and W. Vreuls are with the Department of Pathology, Canisius-Wilhelmina Hospital, 6532SZ Nijmegen, The Netherlands.}
\thanks{S. Mol is with the Department of Pathology, Jeroen Bosch Hospital, 5223GZ 's-Hertogenbosch, The Netherlands.}
\thanks{N. Karssemeijer is with the Diagnostic Image Analysis Group, Radboud University Medical Center, 6500HB Nijmegen, The Netherlands.}
\thanks{This is the author's version of an article that has been published in IEEE Transactions on Medical Imaging. The final version is available at http://dx.doi.org/10.1109/TMI.2018.2820199}
\thanks{Copyright (c) 2018 IEEE. Personal use of this material is permitted. However, permission to use this material for any other purposes must be obtained from the IEEE by sending a request to pubs-permissions@ieee.org.}
}

\maketitle

\begin{abstract}
Manual counting of mitotic tumor cells in tissue sections constitutes one of the strongest prognostic markers for breast cancer. This procedure, however, is time-consuming and error-prone. We developed a method to automatically detect mitotic figures in breast cancer tissue sections based on convolutional neural networks (CNNs). Application of CNNs to hematoxylin and eosin (H\&E) stained histological tissue sections is hampered by: (1) noisy and expensive reference standards established by pathologists, (2) lack of generalization due to staining variation across laboratories, and (3) high computational requirements needed to process gigapixel whole-slide images (WSIs). In this paper, we present a method to train and evaluate CNNs to specifically solve these issues in the context of mitosis detection in breast cancer WSIs. First, by combining image analysis of mitotic activity in phosphohistone-H3 (PHH3) restained slides and registration, we built a reference standard for mitosis detection in entire H\&E WSIs requiring minimal manual annotation effort. Second, we designed a data augmentation strategy that creates diverse and realistic H\&E stain variations by modifying the hematoxylin and eosin color channels directly. Using it during training combined with network ensembling resulted in a stain invariant mitosis detector. Third, we applied knowledge distillation to reduce the computational requirements of the mitosis detection ensemble with a negligible loss of performance. The system was trained in a single-center cohort and evaluated in an independent multicenter cohort from The Cancer Genome Atlas on the three tasks of the Tumor Proliferation Assessment Challenge (TUPAC). We obtained a performance within the top-3 best methods for most of the tasks of the challenge. 
\end{abstract}

\begin{IEEEkeywords}
Breast cancer, mitosis detection, convolutional neural networks, phosphohistone-H3, data augmentation, knowledge distillation
\end{IEEEkeywords}

\IEEEpeerreviewmaketitle

\pagebreak

\section{Introduction}
\begin{figure}[t]
\captionsetup{font=small}
\centering\includegraphics[width=1\linewidth]{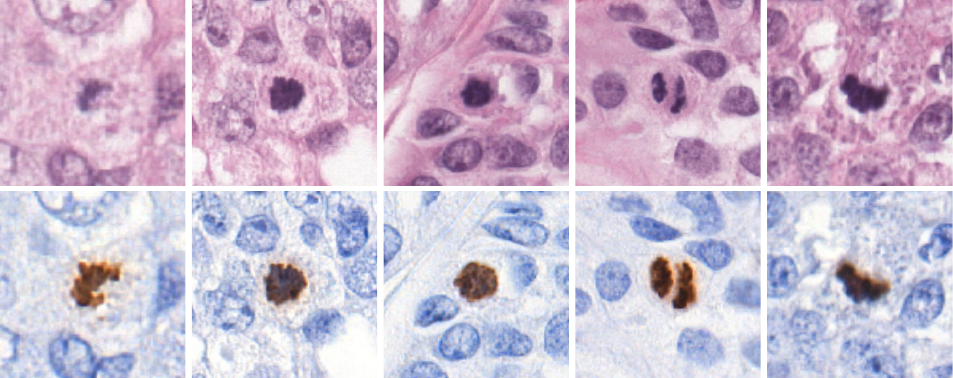}
\caption{Examples of image patches containing mitotic figures, shown at the center of each patch. In H\&E (top), mitotic figures are visible as dark spots. In PHH3 (bottom), they are visible as brown spots. Mitotic figures in PHH3 stain are easier to identify than in H\&E stain.
}
\label{fig:telle1.pdf}
\end{figure}

\IEEEPARstart{H}{istopathological} tumor grade is a strong prognostic marker for the survival of breast cancer patients \cite{bloom1957histological, elston1999pathological}. It is assessed by examination of hematoxylin and eosin (H\&E) stained tissue sections using bright-field microscopy~\cite{van2004prognostic}. Histopathological grading of breast cancer combines information from three morphological features: (1) nuclear pleomorphism, (2) tubule formation and (3) mitotic count, and can take a value within the 1-3 range, where 3 corresponds to the worst patient prognosis. In this study, we focus our attention on the mitosis count component, as it can be used as a reliable and independent prognostic marker \cite{elston1999pathological, skaland2008prognostic}. Mitosis is a crucial phase in the cell cycle where a replicated set of chromosomes is split into two individual cell nuclei. These chromosomes can be recognized in H\&E stained sections as mitotic figures (see Fig. \ref{fig:telle1.pdf}). For breast cancer grading, the counting of mitotic figures is performed by first identifying a region of \SI{2}{\square\mm} with a high number of mitotic figures at low microscope magnification (hotspot) and subsequently counting all mitotic figures in this region at high magnification. 

\begin{figure*}[t]
\captionsetup{font=small}
\centering\includegraphics[width=1\linewidth]{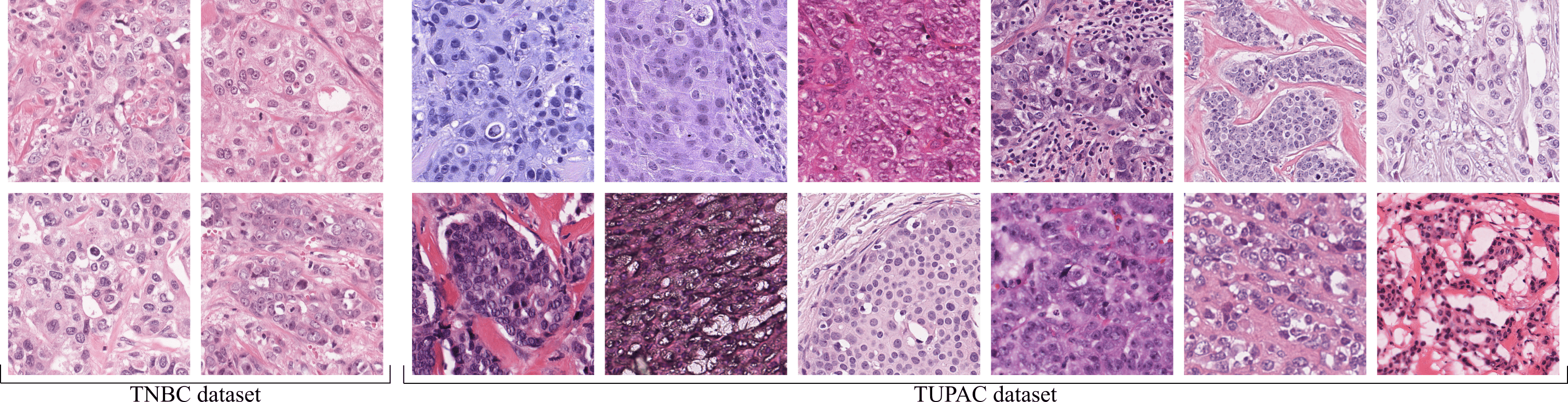}
\caption{Breast tissue samples stained with hematoxylin and eosin (H\&E). Each tile comes from a different patient. The triple negative breast cancer (TNBC) cohort contains images from a single center, whereas the Tumor Proliferation Assessment Challenge (TUPAC) dataset contains images from multiple centers. Notice the homogeneous appearance of the \textit{TNBC} dataset, used for training our mitosis detector, and the variable stain of the \textit{TUPAC} dataset, used to validate our method.}
\label{fig:telle2.png}
\end{figure*}

The recent introduction of whole-slide scanners in anatomic pathology enables pathologists to make their diagnoses on digitized slides \cite{al2013evaluation}{}, so-called whole-slide images (WSIs), and promotes the development of novel image analysis tools for automated and reproducible mitosis counting. Publicly available training datasets for mitosis detection \cite{roux2013mitosis,veta2015assessment,mitos2014,tupac2016} have important limitations in terms of (1) size, (2) tissue representativity, and (3) reference standard agreement. In these datasets, the total number of annotated mitotic figures is currently limited to \num{1500} objects, far away from standard datasets used to train modern computer vision systems \cite{krizhevsky2009learning,deng2009imagenet}. In addition, mitotic figures were annotated in certain manually selected tumor regions only, often excluding tissue areas with image artifacts (common in WSIs). Furthermore, exhaustive manual annotations are known to suffer from disagreement among observers and limited recall \cite{cirecsan2013mitosis,veta2016mitosis}. We propose a method to improve the annotation process based on the automatic analysis of immunohistochemical stained slides. Phosphohistone-H3 (PHH3) is an antibody that identifies cells undergoing mitosis \cite{ribalta2004mitosis,focke2018performance}. Mitotic figures appear in PHH3 immunohistochemically stained slides (abbreviated as PHH3 stained slides) as high contrast objects that are easier to detect than in H\&E \cite{skaland2009validating,colman2006assessment,fukushima2009sensitivity}, illustrated in Fig. \ref{fig:telle1.pdf}. We propose to \emph{destain} H\&E slides and \emph{restain} them with PHH3 to obtain both H\&E and PHH3 WSIs from the exact same tissue section \cite{ikenberg2012immunohistochemical}. By automatically analyzing mitotic activity in PHH3 and registering it to H\&E, we generated training data for mitosis detection in H\&E WSIs in a scalable manner, i.e. independent from the manual annotation procedure.

\begin{table*}[t!]
\captionsetup{font=small}
\caption{Overview of the datasets used in this study. The purpose of \textit{training} indicates that the dataset was used to train a CNN for mitosis detection, whereas \textit{threshold} tuning indicates that it was solely employed to optimize certain hyper-parameters, such as the detection threshold.}
\label{tab:datasets}
\centering
\begin{tabular}{llllll}
\hline
\textbf{Alias}           & \textbf{Data}             & \textbf{Cases} & \textbf{Multicenter} & \textbf{Reference standard}    & \textbf{Purpose} \\ \hline
\textbf{TNBC-H\&E}      & H\&E whole-slide images & 18             & No                    & None                            & Training         \\
\textbf{TNBC-PHH3}      & PHH3 whole-slide images & 18             & No                    & Set of annotated patches                            & Training         \\
\hline
\textbf{TUPAC-train}     & H\&E whole-slide images               & 493            & Yes                   & Grading and proliferation score & Threshold tuning      \\
\textbf{TUPAC-aux-train} & H\&E selected regions      & 50             & Yes                   & Exhaustive location of mitotic figures     & Threshold tuning      \\
\hline
\textbf{TUPAC-test}      & H\&E whole-slide images               & 321            & Yes                   & Not publicly available                            & Evaluation       \\
\textbf{TUPAC-aux-test}  & H\&E selected regions      & 34             & Yes                   & Not publicly available                            & Evaluation       \\
\hline
\end{tabular}
\end{table*}

Although the process of H\&E tissue staining follows a standard protocol, the appearance of the stained slides is not identical among pathology laboratories and varies across time even within the same center (see Fig. \ref{fig:telle2.png}). This variance typically causes mitosis detection algorithms to underperform on images originating from pathology laboratories different than the one that provided the training data \cite{veta2016mitosis}. Several solutions have been proposed to tackle this lack of generalization. First, building multicenter training datasets that contain sufficient stain variability. Following this approach, the Tumor Proliferation Assessment Challenge (TUPAC) resulted in numerous successful mitosis detection algorithms. Top-performing methods in the challenge are based on convolutional neural networks (CNNs)~\cite{cirecsan2013mitosis,Goodfellow-et-al-2016,zerhouni2017wide,veta2016mitosis,paeng2016unified}. This is in line with the trend observed in recent years, which has seen CNNs as the top-performing approach in image analysis, both in computer vision and medical imaging \cite{LITJENS201760}, and corroborates the fact that CNNs have become the standard methodology for automatic mitosis detection. However, multicenter datasets cannot cover all the variability encountered in clinical practice, and are expensive to collect. Second, stain standardization techniques \cite{macenko2009method,khan2014nonlinear,bejnordi2016stain} have been widely used by many of these successful mitosis detection methods to reduce stain variability. However, they require preprocessing all training and testing WSIs and do not reduce the generalization error of trained models. Third, data augmentation strategies have been used to simulate stain variability during the model training. These techniques typically involve RGB transformations such as brightness and contrast enhancements, and color hue perturbations \cite{veta2016mitosis,liu2017detecting}. We argue that designing specific data augmentation strategies for H\&E stained tissue images is the most promising approach to reduce the generalization error of these networks, avoiding the elevated costs of assembling a multicenter cohort, and effectively enforcing stain invariance into the trained models. We propose an augmentation strategy tailored to H\&E WSIs that modifies the hematoxylin and eosin color channels directly, as opposed to RGB, and it is able to generate a broad range of realistic H\&E stain variations from images originating in a single center. We call this technique \emph{stain augmentation}.

Automatic mitosis detection algorithms rely on techniques such as the use of high capacity CNNs and multi-network ensembling to achieve state of the art performance \cite{tupac2016,zerhouni2017wide,cirecsan2013mitosis}. These are simple yet effective mechanisms to improve performance, reduce generalization error and diminish the sensitivity of the model to the detection threshold. However, due to their computationally expensive nature, it is unfeasible to use them for dense prediction in gigapixel WSIs. We propose to exploit the technique of \emph{knowledge distillation} \cite{hinton2015distilling} to reduce the size of the trained ensemble to that of a single network, maintaining similar levels of performance and increasing processing speed drastically.

Our contributions can be summarized as follows:
\begin{itemize}
\item We propose a scalable procedure to exhaustively annotate mitotic figures in H\&E WSIs with minimal human labeling effort. We do so by automatically analyzing mitotic activity in PHH3 restained tissue sections and registering it to H\&E.
\item We propose a data augmentation technique that generates a broad range of realistic H\&E stain variations by modifying the hematoxylin and eosin color channels directly. We demonstrate its ability to enforce stain invariance by transferring the performance obtained in a dataset from a single center to a multicenter publicly available cohort.
\item We apply knowledge distillation to reduce the size of an ensemble of trained networks to that of a single network, in order to perform mitosis detection in gigapixel WSIs with similar performance and vastly increased processing speed. 
\end{itemize}

\begin{figure*}[t!]
\captionsetup{font=small}
\centering\includegraphics[width=0.9\linewidth]{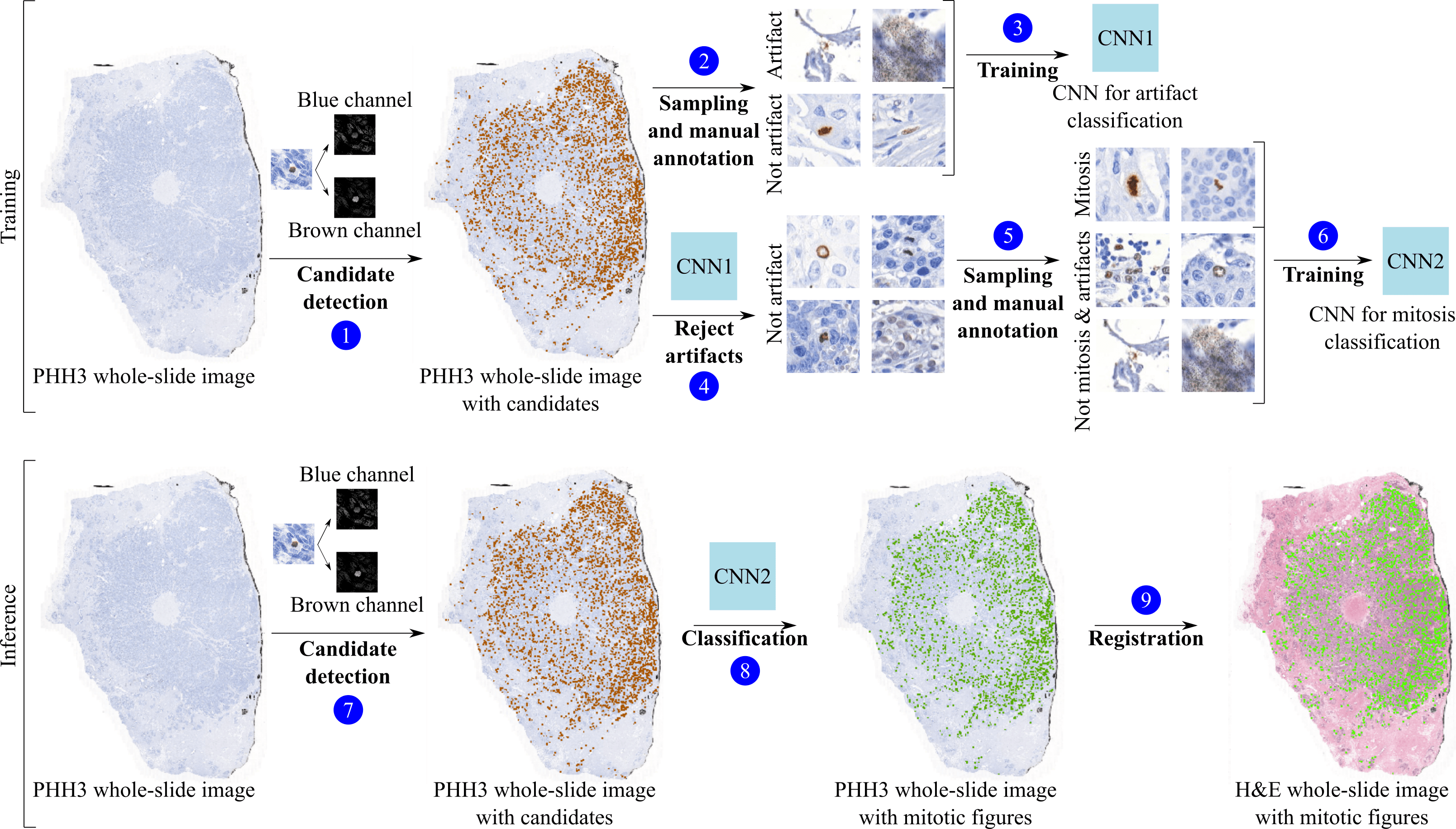}
\caption{Building reference standard for mitotic activity using PHH3 stained slides. Top: training stage, where mitotic candidates are extracted from the brown color channel (1), pruned from artifacts (2, 3, 4), a subset is manually annotated (5), and then used to train a CNN to distinguish between mitotic and non-mitotic patches (6), named \textit{CNN2}. Bottom: inference stage, where candidates in a given PHH3 slide (7) are classified with \textit{CNN2} as mitotic or non-mitotic (8), then registered to their respective H\&E slide pairs (9).}
\label{fig:telle3.png}
\end{figure*}

The paper is organized as follows. Sec. \ref{sec:materials} reports the datasets used to train and validate our method. Sec. \ref{sec:phh3} and Sec. \ref{sec:he} describe the methodology in depth. All details regarding CNN architectures, training protocols and hyper-parameter tuning are explained in Sec. \ref{sec:net}. Experimental results are listed in Sec. \ref{sec:exp_res}, followed by Sec. \ref{sec:discussion} where the discussion and final conclusion are stated.

\section{Materials}
\label{sec:materials}

In this study, we use two cohorts from different sources for (1) developing the main mitosis detection algorithm, and (2) performing an independent evaluation of the system performance. Details on the datasets are provided in Table \ref{tab:datasets}. 

The first cohort consists of 18 triple negative breast cancer (TNBC) patients who underwent surgery in three different hospitals in the Netherlands: Jeroen Bosch Hospital, Radboud University Medical Centre (Radboudumc) and Canisius-Wilhelmina Hospital. However, all tissue sections were cut, stained and scanned at the Radboudumc using a 3DHistech Pannoramic 250 Flash II scanner at a spatial resolution of \SI{0.25}{\um/pixel}, therefore, we consider this set of WSIs a single-center one. Subsequently, the slides were destained, restained with PHH3 and re-scanned, resulting in 18 pairs of H\&E and PHH3 WSIs representing the exact same tissue section per pair. We will refer to these images as the \textit{TNBC-H\&E} and \textit{TNBC-PHH3} datasets through the rest of the paper. 

The second cohort consists of the publicly available TUPAC dataset~\cite{tupac2016}. In particular, 814 H\&E WSIs from invasive breast cancer patients from multiple centers included in The Cancer Genome Atlas \cite{weinstein2013cancer} scanned at \SI{0.25}{\um/pixel} were annotated, providing two labels for each case. The first label is the histological grading of each tumor based on the mitotic count only. The second score is the outcome of a molecular test highly correlated with tumor proliferation \cite{nielsen2010comparison}. Out of these 814 WSIs, 493 cases have a public reference standard, whereas the remaining 321 cases do not (only available to TUPAC organizers). We will refer to these sets of WSIs as the \textit{TUPAC-train} and \textit{TUPAC-test} datasets respectively. 

Additionally, the organizers of TUPAC provided data for individual mitotic figure detection from two centers. They preselected 84 proliferative tumor regions from H\&E WSIs of breast cancer cases, and two observers exhaustively annotated them. Coincident annotations were marked as the reference standard, and discording items were reviewed by a third pathologist. Out of these 84 regions, 50 cases have a public reference standard, whereas the remaining 34 cases do not (only available to TUPAC organizers). We will refer to these sets as the \textit{TUPAC-aux-train} and \textit{TUPAC-aux-test} datasets respectively. 

All WSIs in this work were preprocessed with a tissue-background segmentation algorithm \cite{bandi2017comparison} in order to exclude areas not containing tissue from the analysis.

\section{PHH3 stain: reference standard for \\mitotic activity}
\label{sec:phh3}

We trained a CNN to automatically identify mitotic figures throughout PHH3 WSIs and registered the detections to the H\&E WSI pairs. The entire process is summarized in Fig.~\ref{fig:telle3.png}. 

\subsection{Mitosis detection in PHH3 whole-slide images}

We used color deconvolution to disentangle the DAB and hematoxylin stains (brown and blue color channels respectively) and obtained the mitotic candidates by labeling connected components of all positive pixels belonging to the DAB stain. We observed that the PHH3 antibody was sensitive but not specific regarding mitosis activity: 75\% of candidates were trivial artifacts. To avoid waste of manual annotation effort on these artifacts, we trained a CNN, named \textit{CNN1}, to classify candidates among artifactual or non-artifactual objects. To train such a system, a student labeled 2000 randomly selected candidates. We classified all PHH3 candidates with \textit{CNN1}, and randomly selected 2000 samples classified as non-artifactual. Four observers labeled these samples as either containing a mitotic figure or not. These four observers consisted of a pathology resident, a PhD student and two lab technicians, and they were provided with sufficient training, visual examples and hands-on practice on mitosis detection. Annotations were aggregated by performing majority voting, keeping only those samples where at least 3 observers agreed upon. This resulted in 778 and 1093 mitotic and non-mitotic annotations, respectively. Furthermore, the non-mitotic set was extended with 1500 artifactual samples used to train \textit{CNN1}, resulting in 2593 non-mitotic samples.

We used this annotated set of samples to train \textit{CNN2} to distinguish PHH3 candidates among mitotic and non-mitotic patches. During training, we randomly applied several techniques to augment the data and prevent overfitting, namely: rotations, vertical and horizontal mirroring, elastic deformation \cite{simard2003best}, Gaussian blurring, and translations. The resulting performance of \textit{CNN2} was an F1-score of 0.9. Details on network architecture, training protocol and hyper-parameter selection are provided in Sec. \ref{sec:net}. We classified all candidates found in the PHH3 slides as mitotic or non-mitotic objects using \textit{CNN2}, generating exhaustive reference standard data for mitosis activity at whole-slide level.

\subsection{Registering mitotic figures from PHH3 to H\&E slides}

The process of slide restaining guaranteed that the exact same tissue section was present in both the H\&E and the PHH3 WSIs, requiring minimal registration to align mitotic objects. We designed a simple yet effective two-step routine to reduce vertical and horizontal shift at \emph{global} and \emph{local} scale. First, we performed a global and coarse registration that minimized the vertical and horizontal shift between image pairs. We did so by finding the alignment of randomly selected pairs of corresponding tiles as the shift vector that maximized the 2D cross-correlation. For improved accuracy, we repeated this procedure 10 times per WSI pair (at random locations throughout the WSI), averaging the cross-correlation heatmap across trials. Finally, all mitoses were adjusted with the resulting global shift vector. Second, we registered each mitotic figure individually, following a similar procedure as before. We extracted a single pair of high magnification tiles, centered in each candidate location, to account for individual local shifts.

\begin{figure*}[ht!]
\captionsetup{font=small}
\centering\includegraphics[width=1\linewidth]{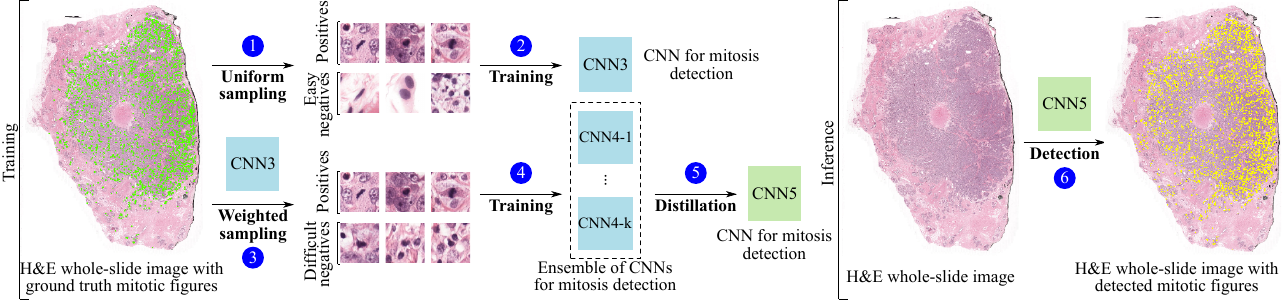}
\caption{
Training a whole-slide H\&E mitosis detector. Left (training stage): an auxiliary network \textit{CNN3} is trained with uniformly sampled patches (1, 2). Then, \textit{CNN3} is used to perform hard mining on the negative patches and create \textit{k} distinct datasets sampled via bootstrapping (3). These datasets are used to train networks \textit{CNN4-1} to \textit{CNN4-k} and build an ensemble (4). Finally, the ensemble is reduced into \textit{CNN5} via knowledge distillation (5). Right (inference stage): \textit{CNN5} is applied in a fully convolutional manner throughout an entire WSI to detect mitotic figures exhaustively (6).
}
\label{fig:telle4.pdf}
\end{figure*}

\begin{figure}[t]
\captionsetup{font=small}
\centering\includegraphics[width=1\linewidth]{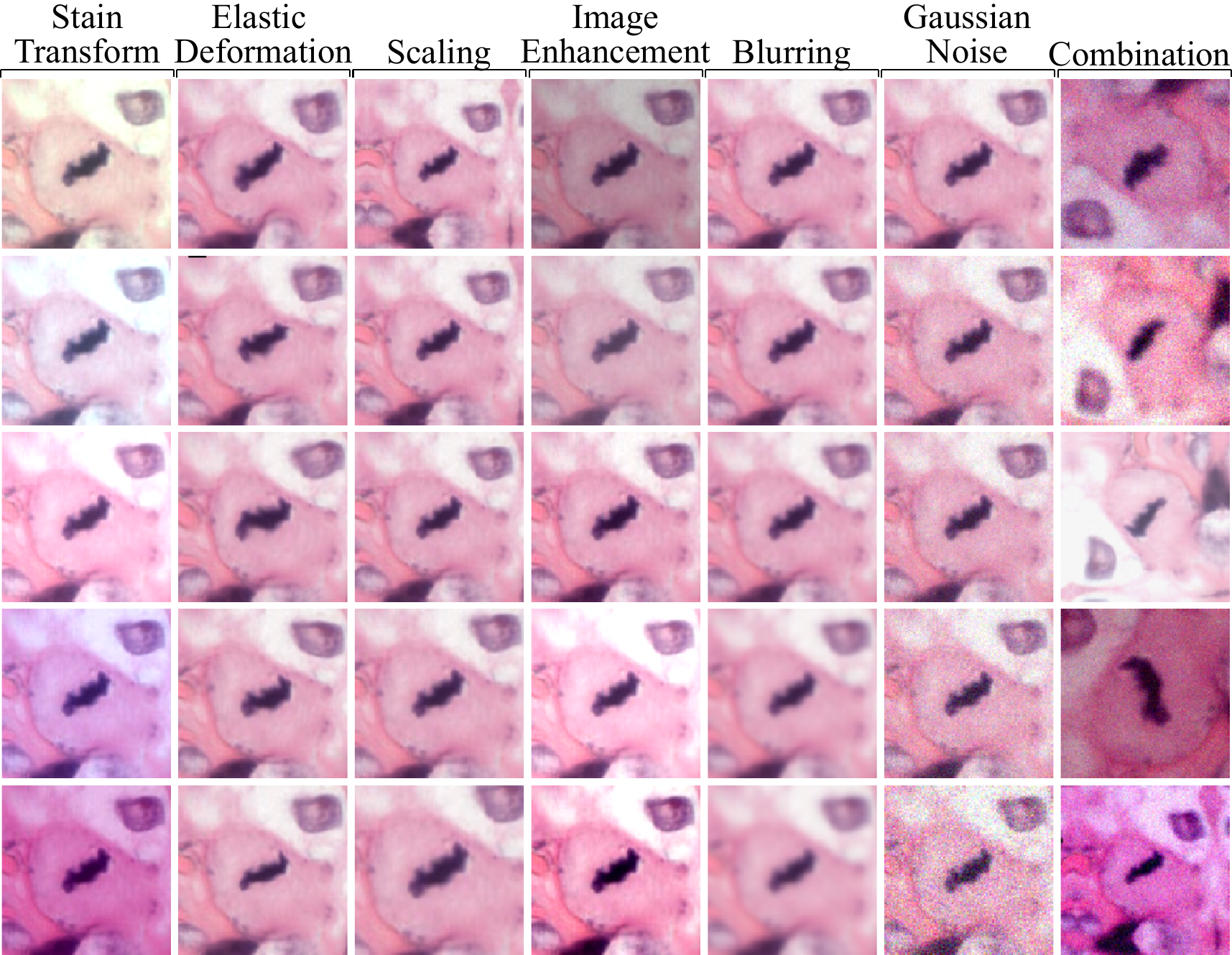}
\caption{Multiple augmented versions of the same mitotic patch. Each column shows samples of a single augmentation function except for the last one, which combines all the techniques together with rotation and mirroring.}
\label{fig:patches_augmented}
\end{figure}
 
\section{H\&E stain: training a mitosis detector}
\label{sec:he}

We trained a CNN for the task of mitosis detection and used it to exhaustively locate mitotic figures throughout H\&E WSIs. Only slides from the \textit{TNBC-H\&E} dataset were used in this step. The procedure is summarized in Fig. \ref{fig:telle4.pdf}.

\subsection{Assembling a training dataset for mitosis detection}

Even though the \textit{TNBC-H\&E} dataset already possessed little stain variation, we standardized the stain of each WSI to reduce intra-laboratory stain variations \cite{bejnordi2016stain}, preventing the CNN from becoming stain invariant from the raw training data. This further strengthens the challenge of generalizing to unseen stain variations.

As a result of the large amount of available pixels in WSIs, the selection of negative samples was not trivial. We propose a \emph{candidate detector} based on the assumption that mitotic figures are non-overlapping circular objects with dark inner cores. This detector found candidates by iterative detection and suppression of all circular areas of diameter $d$ centered on local minima of the mean RGB intensity until all pixels above a threshold $t$ were exhausted. We selected a sufficiently large $t$ so that candidates were representative for all tissue types. A candidate was labeled as a positive sample if its Euclidean distance to any reference standard object was at most $d$ pixels, and labeled as a negative sample otherwise. 

Most of the negative samples were very easy to classify and their contribution to improve the decision boundaries of the CNN was marginal. We found it crucial to identify highly informative negative samples to train the CNN effectively. We proceeded similarly as stated in \cite{cirecsan2013mitosis}. First, we built an \textit{easy} training dataset by including all positive candidates and a number of uniformly sampled negative candidates, and trained a network with it, labeled as \textit{CNN3} for future reference. Second, we evaluated all candidate patches with this network, obtaining a prediction probability for each of them. Finally, we built a \textit{difficult} training dataset by selecting all positive candidates, and a number of negative candidates sampled proportionally to their probability of being mitosis, so that harder samples were chosen more often. 

\begin{figure}[t]
\captionsetup{font=small}
\centering\includegraphics[width=1\linewidth]{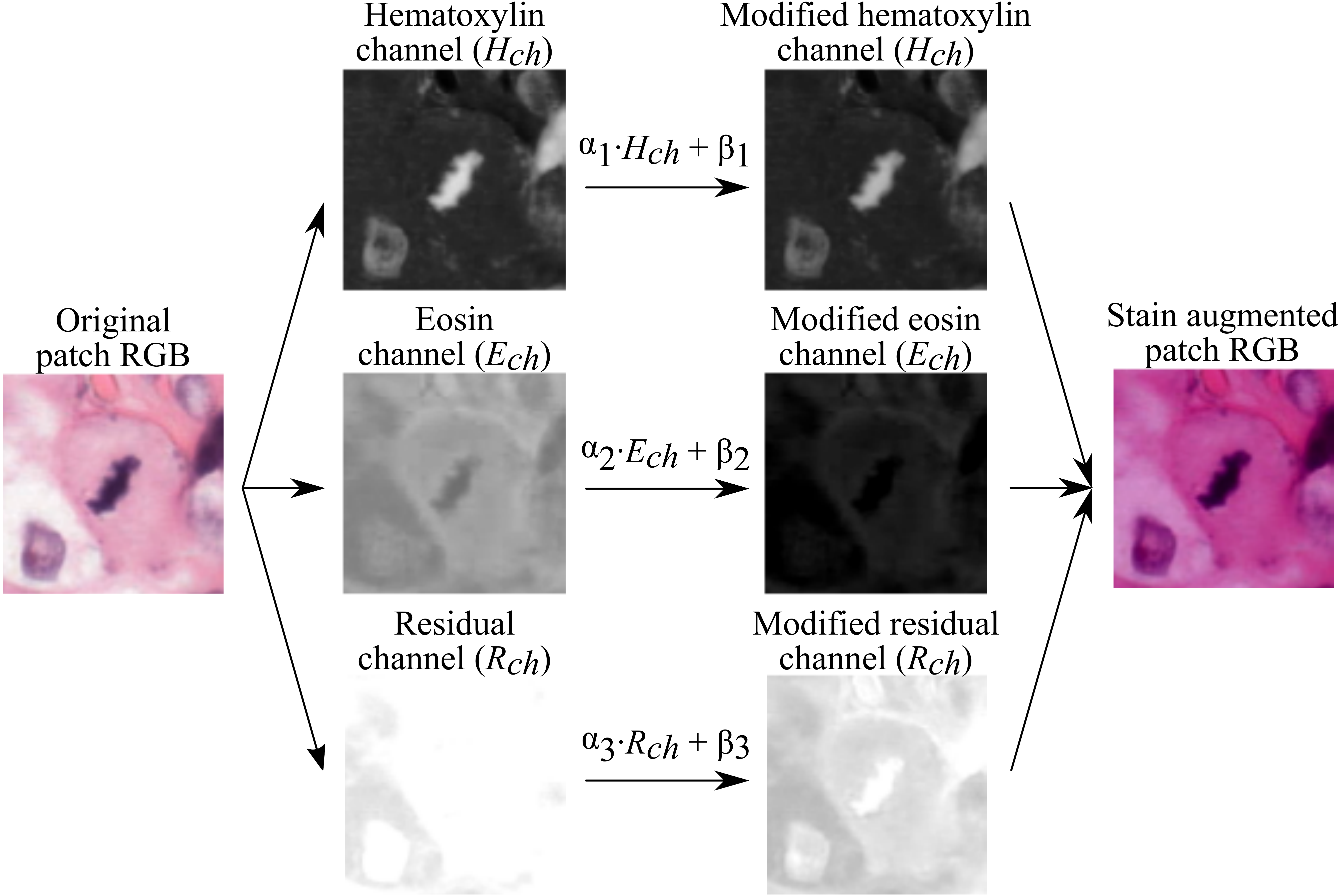}
\caption{H\&E stain augmentation. From left to right: first, an RGB patch is decomposed into hematoxylin ($H_{ch}$), eosin ($E_{ch}$) and residual ($R_{ch}$) color channels. Then, each channel is individually modified with a random factor and bias. Finally, resulting channels are transformed back to RGB color space.}
\label{fig:telle6.pdf}
\end{figure}

\subsection{H\&E stain augmentation}

We trained a CNN on the \textit{difficult} dataset to effectively distinguish between mitotic and non-mitotic H\&E patches, named as \textit{CNN4}. During training, we applied several techniques to augment the dataset on-the-fly, preventing overfitting and improving generalization. We implemented several routines for H\&E histopathology imaging in the context of mitosis detection, illustrated in Figure \ref{fig:patches_augmented} with a sample patch. 

\textbf{Morphology invariance}. We exploited the fact that mitotic figures can have variable shapes and sizes by augmenting the training patches with rotation, vertical and horizontal mirroring (\textit{R}), scaling (\textit{S}), elastic deformation (\textit{E}) \cite{simard2003best}, and translation around the central pixel. 

\textbf{Stain invariance}. We used a novel approach to simulate a broad range of realistic H\&E stained images by retrieving and modifying the intensity of the hematoxylin and eosin color channels directly (\textit{C}), as illustrated in Figure \ref{fig:telle6.pdf}. First, we transformed each patch sample from RGB to H\&E color space using a method based on color deconvolution \cite{ruifrok2001quantification}{}, see the Appendix for more methodological details. Second, we modified each channel $i$ individually, i.e. hematoxylin ($H_{ch}$), eosin ($E_{ch}$) and residual ($R_{ch}$), with random factors $\alpha_i$ and biases $\beta_i$ taken from two uniform distributions. Finally, we combined and projected the resulting color channels back to RGB. Additionally, we simulated further alternative stains by modifying image brightness, contrast and color intensity (\textit{H}). 

\textbf{Artifact invariance}. We mimicked the out of focus effect of whole-slide scanners with a Gaussian filter  (\textit{B}), and added Gaussian noise to decrease the signal-to-noise ratio of the images (\textit{G}), simulating image compression artifacts. 

\subsection{Ensemble \& network distillation}

The use of an ensemble of networks is a key factor to achieve state of the art performance in multiple classification tasks \cite{deng2009imagenet}, particularly in mitosis detection \cite{cirecsan2013mitosis,zerhouni2017wide}. An ensemble of networks performs significantly better than any of its members if they make independent (uncorrelated) errors. Building different training datasets with replacement (bagging) has been shown to increase model independence in an ensemble \cite{Goodfellow-et-al-2016}. Therefore, we trained \textit{k} different CNNs on \textit{k} different training datasets, each one obtained by sampling negative candidates with replacement, and made an ensemble with the networks, averaging their predicted probabilities across models.

The computational requirements of this ensemble grow proportionally to \textit{k}. To reduce this burden, we applied the idea of knowledge distillation, a technique designed to transfer the performance of a network (or ensemble of networks) to a lighter target neural network \cite{hinton2015distilling}{}. To achieve the highest performance, we distilled the ensemble of \textit{k} networks to a single smaller CNN, named \textit{CNN5}. We did so by training \textit{CNN5} directly on the continuous averaged output probabilities of the ensemble, instead of the dataset labels, as indicated in~\cite{hinton2015distilling}{}. We defined $\gamma$ as a parameter to control the amount of trainable parameters used by \textit{CNN5}, taking values in the $[0, 1]$ range. In particular, the number of filters per convolutional layer was proportional to this parameter. It defaults to $\gamma=1$, unless stated otherwise.

\subsection{Outcome at whole-slide level}

We detected mitotic figures at whole-slide level by sliding \textit{CNN5} over tissue regions at \SI{0.25}{\um/pixel} resolution, producing a mitosis probability map image for each slide. Simple post-processing allowed to detect individual mitotic objects: (1) thresholding and binarizing the probability map (pixel probability of at least 0.8 to reduce the computational burden), (2) labeling connected components in the resulting image, and (3) suppressing double detections closer than $d$ pixels. A detection probability threshold $\delta$ must be provided to discern between false and true positives.

In order to provide the number of mitotic figures in the most active tumor area per slide, we slid a virtual hotspot consisting of a \SI{2}{\square\mm} circle throughout each entire WSI, counting the number of mitoses at each unique spatial position. To identify the hotpot with the largest mitotic activity ignoring large outliers, we considered the 95th percentile of the series of mitotic countings for each slide, excluding empty areas.

To estimate the tumor grading of the patient, we followed the guidelines used to compute the Mitotic Activity Index (MAI) \cite{van2004prognostic}. We defined two thresholds, $\theta_1$ and $\theta_2$, and used them to categorize the number of mitotic figures in the hotspot into three possible outcomes. In particular, we predicted grade 1, 2 or 3 depending on whether the mitotic counting was below or equal to $\theta_1$, between thresholds, or above $\theta_2$, respectively. To estimate a continuous tumor proliferation score, we simply provided the number of mitotic figures in the hotspot.

\begin{table}[t]
\captionsetup{font=small}
\caption{Architecture of \textit{CNN3}, \textit{CNN4} and \textit{CNN5}. $\gamma$ controls the number of filters per convolutional layer. }
\label{tab:network-architecture}
\centering
\begin{tabular}{llllll}
\hline
 \textbf{Function} & \textbf{Filters} & \textbf{Size} & \textbf{Stride} & \textbf{Activation} \\
\hline
conv     & 32 $\gamma$    & 3x3         & 1      & Leaky-ReLU \\
conv     & 32 $\gamma$    & 3x3         & 2      & Leaky-ReLU \\
conv     & 64 $\gamma$    & 3x3         & 1      & Leaky-ReLU \\
conv     & 64 $\gamma$    & 3x3         & 2      & Leaky-ReLU \\
conv     & 128 $\gamma$   & 3x3         & 1      & Leaky-ReLU \\
conv     & 128 $\gamma$   & 3x3         & 1      & Leaky-ReLU \\
conv     & 256 $\gamma$   & 3x3         & 1      & Leaky-ReLU \\
conv     & 256 $\gamma$   & 3x3         & 1      & Leaky-ReLU \\
conv     & 512 $\gamma$   & 14x14       & 1      & Leaky-ReLU \\
dropout  & -       & -           & -      & -          \\
conv     & 2       & 1           & 1      & Softmax  \\
\hline
\end{tabular}
\end{table}

\section{CNN architecture, training protocol and other hyper-parameters}
\label{sec:net}

In order to train the CNN models, we used RGB patches of 128$\times$128 pixel size, taken at \SI{0.25}{\um/pixel} resolution and whose central pixel was centered in the coordinates of the annotated object. Patches were cropped as part of the data augmentation strategy, resulting in 100$\times$100 pixel images fed into the CNNs. 

Convolutional neural networks were trained to minimize the cross-entropy loss of the network outputs with respect to the patch labels, using stochastic gradient descent with Adam optimization and balanced mini-batch of 64 samples. To prevent overfitting, an additional $L_2$ term was added to the network loss, with a factor of \SI{1e-5}{}. Furthermore, the learning rate was exponentially decreased from \SI{1e-3} to \SI{3e-5} through 20 epochs. At the end of training, network parameters corresponding to the checkpoint with the highest validation F1-score were selected for inference.

\subsection{Mitosis detection in PHH3}

\textit{CNN1} was trained with 1500 artifactual and 500 non-artifactual samples, and \textit{CNN2} was trained with 778 mitotic and 2593 non-mitotic patch samples. In both cases, the sets of samples were randomly divided into training and validation subsets at case level from \textit{TNBC-PHH3}, with 10 and 8 slides for training and validation, respectively. The architecture of \textit{CNN1} and \textit{CNN2} consisted of five pairs of convolutional and max-pooling layers with 64, 128, 256, 512 and 1024 3$\times$3 filters per layer, followed by two densely connected layers of 2048 and 2 units respectively. A dropout layer was placed between the last two layers, with 0.5 coefficient. All convolutional and dense layers were followed by ReLU functions, except for the last layer that ended with a softmax function. 

\begin{table}[t]
\captionsetup{font=small}
\caption{Analysis of the impact in performance of using data augmentation, ensemble and knowledge distillation. Each row represents a CNN (or set of CNNs for the ensemble case) that was trained using \textit{TNBC-H\&E} data as explained in Sec. IV. Each trained network was evaluated in the independent \textit{TUPAC-aux-train} dataset with multiple detection thresholds, reporting the highest F1-score obtained and the number of trainable parameters. Experiments 1, 2 and 3 compared the use of different data augmentation strategies (R: rotation, C: color stain, S: scaling, etc., see Sec. IV.B for the full list). Experiment 4 showed the performance of an ensemble of $k=10$ networks, trained as explained in Sec. IV.C. In experiments 5, 6 and 7, the ensemble of CNNs (experiment 4) was distilled into single smaller CNNs with varying capacities $\gamma=1.0, 0.8, 0.6$ as explained in Sec. IV.C. The CNN trained for experiment 7 coincides with \textit{CNN5}.
}
\label{tab:aug}
\centering
\begin{tabular}{llllll}
\hline
\textbf{Exp} & \textbf{Augment} & \textbf{Ensemble} & \textbf{Distilled} & \textbf{F1-score} & \textbf{Param} \\
\hline
1 & RSEB         & No          & No        & 0.018 & 26.9M     \\
2 & RCSEB        & No          & No        & 0.412 & 26.9M     \\
3  & RCSEHBG      & No          & No       & 0.613 & 26.9M     \\
 \hline
4 & RCSEHBG      & $k=10$         & No        & 0.660  & 269M    \\
 \hline
5 & RCSEHBG      & No          & $\gamma=1.0$      & 0.623  & 26.9M     \\
6 & RCSEHBG      & No          & $\gamma=0.8$      & 0.628 & 17.1M       \\
7 & RCSEHBG      & No          & $\gamma=0.6$      & 0.636 & 9.5M    \\
\hline
\end{tabular}
\end{table}

\subsection{Mitosis detection in H\&E}

The H\&E slides provided in the \textit{TNBC-H\&E} dataset were randomly divided into training, validation and test subsets, with 11, 3 and 4 slides each. For the candidate detector, we selected $d=100$ as the diameter of an average tumor cell at \SI{0.25}{\um/pixel} resolution; and $t=0.6$ recalling $99\%$ of the mitotic figures in the reference standard of the validation set, with a rate of 1 positive to every 1000 negative samples. On average, each slide had 1 million candidates. The architecture of \textit{CNN3}, \textit{CNN4} and \textit{CNN5} is summarized in Table \ref{tab:network-architecture}. We found that substituting max-pooling layers for strided convolutions slightly improved convergence, resulting in an \textit{all convolutional} architecture \cite{springenberg2014striving}. To train \textit{CNN3}, we built a training set with all positive samples, \num{22000} mitotic figures, and \num{100000} uniformly sampled negative candidates. For validation purposes, we also built a validation set consisting of 10\% of the total available samples in both classes, \num{200000} negative and 500 positive candidates. To train \textit{CNN4} and \textit{CNN5}, we built a training set with all positive samples, \num{22000} mitotic figures, and \num{400000} negative candidates, sampling difficult patches more often with replacement. We used the same validation set as with \textit{CNN3}. For the ensemble, we selected $k=10$, the highest number of networks that we could manage in an ensemble with our computational resources. We distilled several versions of \textit{CNN5} varying the value of $\gamma$, evaluated each network in the validation set of the \textit{TNBC-H\&E} dataset, and obtained an F1-score of 0.634, 0.646 and 0.629 for $\gamma$ values of 1.0, 0.8 and 0.6, respectively. We selected $\gamma=0.6$ for further experiments, resulting in a distilled network with 28X and 2.8X times less parameters than the ensemble and the single network ($\gamma=1.0$), respectively, at a negligible cost of performance.

The color augmentation technique sampled $\alpha$ and $\beta$ from two uniform distributions with ranges [0.95, 1.05] and [-0.05, 0.05], respectively. Patches were scaled with a zooming factor uniformly sampled from [0.75, 1.25]. The elastic deformation routine used $\alpha=100$, and $\sigma=10$. Color, contrast and brightness intensity was enhanced by factors uniformly sampled from [0.75, 1.5], [0.75, 1.5] and [0.75, 1.25], respectively. The Gaussian filter used for blurring sampled $\sigma$ uniformly from the [0, 2] range. The additive Gaussian noise had zero mean and a standard deviation uniformly sampled from the [0, 0.1] range. These parameters were selected empirically to simultaneously maximize visual variety and result in realistic samples.

\section{Experiments and results}
\label{sec:exp_res}

\begin{figure}[t]
\captionsetup{font=small}
\centering\includegraphics[width=1\linewidth]{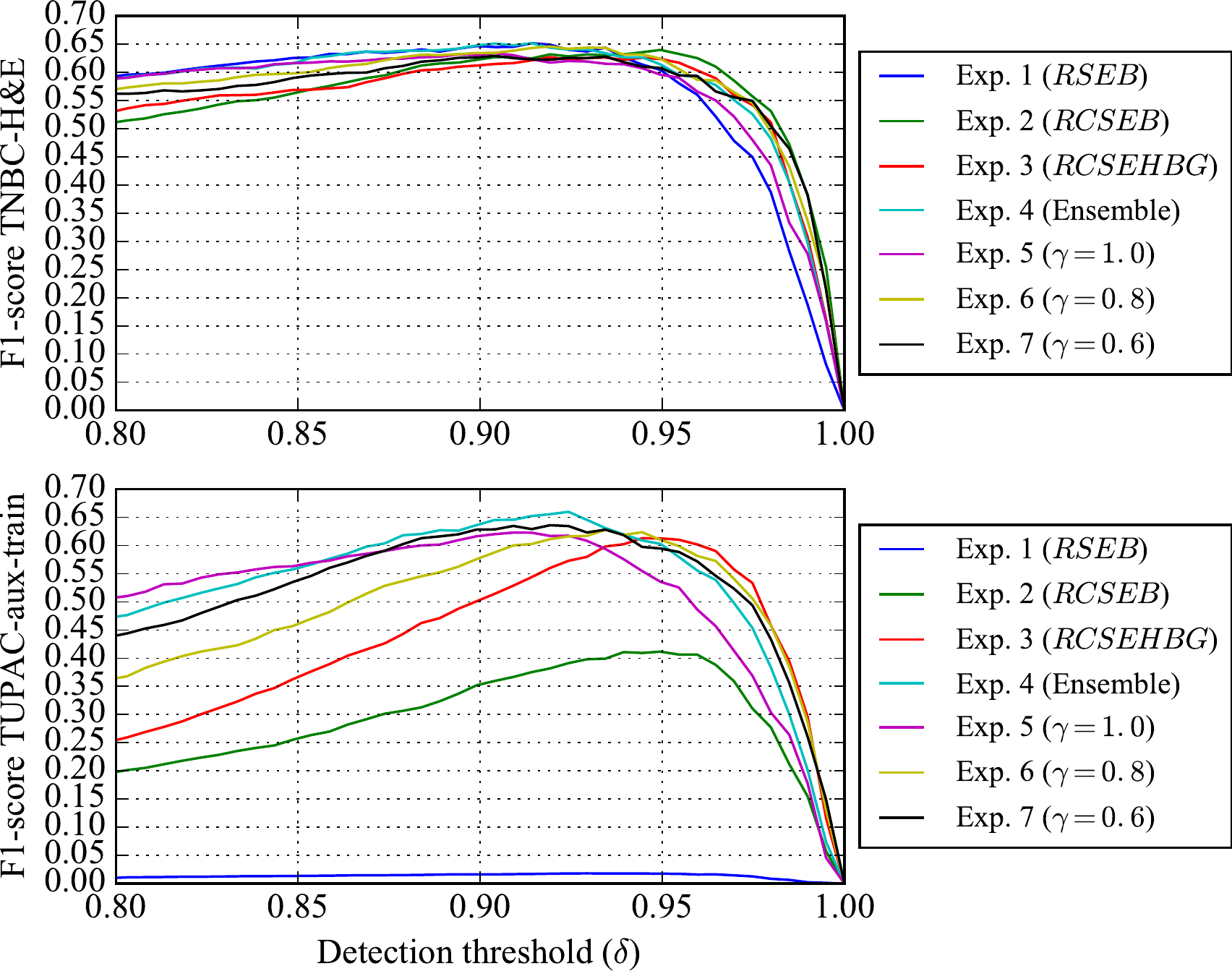}
\caption{Analysis of the impact in performance of using data augmentation, ensemble and knowledge distillation measured in terms of F1-score with respect to the detection threshold (top: \textit{TNBC-H\&E dataset}, bottom: \textit{TUPAC-aux-train dataset}). Experiments 1, 2 and 3 compared the use of different data augmentation strategies (R: rotation, C: color stain, S: scaling, etc., see Sec. IV.B for the full list). Experiment 4 showed the performance of an ensemble of $k=10$ networks, trained as explained in Sec. IV.C. In experiments 5, 6 and 7, the ensemble of CNNs (experiment 4) was distilled into single smaller CNNs with varying capacities $\gamma=1.0, 0.8, 0.6$ as explained in Sec. IV.C. The CNN trained for experiment 7 coincides with \textit{CNN5}.}
\label{fig:telle7.pdf}
\end{figure}

\begin{table*}[ht]
\captionsetup{font=small}
\caption{Independent evaluation of the proposed method performance in the three tasks of the TUPAC challenge. Columns Top-1, Top-2 and Top-3 correspond to the best performing solutions in the public leaderboard, respectively. }
\label{tab:exp_res}
\centering
\begin{tabular}{lllllll}
\hline
\textbf{Dataset}   & \textbf{Ground truth} & \textbf{Metric}  & \textbf{Top-1} & \textbf{Top-2} & \textbf{Top-3}  & \textbf{Proposed [95 c.i.]} \\
\hline
TUPAC-test     & Tumor grading     & Kappa    & 0.567 & 0.534 & 0.462    & 0.471 [0.340, 0.603] \\
TUPAC-test    & Proliferation score   & Spearman   &  0.617 & 0.516 & 0.503    & 0.519 [0.477, 0.559] \\
TUPAC-aux-test & Mitosis location   & F1-score & 0.652 & 0.648 & 0.616    & 0.480    \\
\hline 
\end{tabular}
\end{table*}

\subsection{Impact of augmentation, ensembling and distillation}

We performed a series of experiments to quantitatively assess the impact in performance of three ideas mentioned in this paper: (a) data augmentation, (b) ensemble and (c) knowledge distillation. In each experiment, we trained a CNN (or set of CNNs for the ensemble case) with the \textit{TNBC-H\&E} dataset as explained in Sec. IV and V. Each trained model was evaluated in the independent \textit{TUPAC-aux-train} dataset with multiple detection thresholds, reporting the highest F1-score obtained. Table \ref{tab:aug} summarizes the numerical results, and Figure \ref{fig:telle7.pdf} analyzes the sensitivity of each model with respect to the detection threshold.

\textbf{Data augmentation}. The goal of experiments 1, 2 and 3 is to test whether the proposed data augmentation strategy can improve the performance of the CNN in an independent test set, in particular the novel color stain augmentation. In experiment 1, we trained a baseline system including only basic augmentation (\textit{RSEB}) and obtained an F1-score of 0.018. In experiment 2, we repeated the training procedure including our color augmentation technique as well (\textit{RCSEB}) and obtained an F1-score of 0.412. In experiment 3, we repeated the training procedure including all the augmentation techniques mentioned in Sec. IV.B (\textit{RCSEHBG}) and obtained an F1-score of 0.613.

\textbf{Ensemble}. The goal of experiment 4 is to test whether the use of an ensemble of networks can improve the performance of the mitosis detector beyond the results obtained in experiment 3 with a single network. We trained and combined a set of CNNs, as explained in Sec. IV.C, and obtained an F1-score of 0.660. Furthermore, we analyzed its performance with respect to the detection threshold and observed a more robust behavior than that of the single CNN tested in experiment 3, illustrated in Fig. 7.

\textbf{Distillation}. The goal of experiments 5, 6 and 7 is to test whether knowledge distillation can effectively transfer the performance of the ensemble trained in experiment 4 to a single CNN. We trained three CNNs with $\gamma$ set to 1.0, 0.8 and 0.6, respectively to experiments 5, 6 and 7. They all exhibited a similar performance to that of the ensemble (F1-score of 0.623, 0.628 and 0.636, respectively), with drastically less trainable parameters and superior performance to a single CNN trained without distillation (experiment 3). Experiment 7 resulted in \textit{CNN5}, used in the following sections.

\subsection{Comparison with the state of the art}

We evaluated the performance of our system in the three tasks of the TUPAC Challenge \cite{tupac2016}, and compared the results with those of top-performing teams, summarized in Tab. \ref{tab:exp_res}. We used \textit{CNN5} with $\gamma=0.6$ for all submissions, solely trained with the \textit{TNBC-H\&E} dataset. Notice that the authors do not have access to the ground truth data of \textit{TUPAC-test} and \textit{TUPAC-aux-test}. Our model predictions were independently evaluated by the organizers of TUPAC. This ensured fair and independent comparison with state of the art methods. For the first and second tasks, we tuned the hyper-parameters of the proposed whole-slide mitosis detector with the \textit{TUPAC-train} dataset. We selected $\delta=0.970$ to maximize the Spearman correlation between our mitotic count prediction and the ground truth proliferation score. Then, we tuned $\theta_1$ and $\theta_2$ to maximize the quadratic weighted Cohen's kappa coefficient between our predicted tumor grade and the ground truth, obtaining $\theta_1=6$ and $\theta_2=20$. For the third task, we selected $\delta=0.919$ to maximize the F1-score metric in the \textit{TUPAC-aux-train} dataset. Spearman, kappa and F1-score are the evaluation metrics proposed in the TUPAC Challenge.

For the first task, we obtained a Kappa agreement of 0.471 with 95 confidence intervals [0.340, 0.603] between the ground truth tumor grading and our prediction on the \textit{TUPAC-test} dataset. This performance is comparable to the top-3 entry in the leaderboard.

For the second task, we obtained a Spearman correlation of 0.519 with 95 confidence intervals [0.477, 0.559] between the ground truth genetic-based proliferation score and our prediction on the \textit{TUPAC-test} dataset. This performance is comparable to the top-2 entry in the leaderboard.

For the third task, we obtained an F1-score of 0.480, with a precision and recall values of 0.467 and 0.494, respectively, by detecting individual mitotic figures in the \textit{TUPAC-aux-test}. This performance is comparable to the top-7 entry in the leaderboard.

\subsection{Observer study: precision of the mitosis detector}

Due to the relatively low F1-score of 0.480 obtained in the \textit{TUPAC-aux-test}, compared to the F1-score of 0.636 obtained in the \textit{TUPAC-aux-train}, we investigated whether this difference could potentially be caused by a combination of inter-observer variability in the TUPAC reference standard, which was established by human observers, and lack of sufficient number of test samples. A resident pathologist manually classified the detections of \textit{CNN5} on the \textit{TUPAC-aux-test}, blinded to the patch labels. The observer indicated that 128 out of 181 detections contained mitotic figures, resulting in a precision of 0.707 for the detector. With this precision and assuming the recall suggested by the organizers, we would obtain an alternative F1-score of 0.581 in the \textit{TUPAC-aux-test}. For the sake of completeness, the patches used in this experiment are depicted in the Appendix (Fig. \ref{fig:telle8.pdf}). 

\section{Discussion and Conclusion}
\label{sec:discussion}

To the best of our knowledge, this is the first time that the problem of noisy reference standards in training algorithms for mitosis detection in H\&E WSIs was solved using immunohistochemistry. We validated our hypothesis that mitotic activity in PHH3 can be exploited to train a mitosis detector for H\&E WSIs that is competitive with the state of the art. We proposed a method that combined (1) H\&E-PHH3 restaining, (2) automatic image analysis and (3) registration to exhaustively annotate mitotic figures in entire H\&E WSIs, for the first time. Only 2 hours of manual annotations per observer were needed to train the algorithm, delivering a dataset that was at least an order of magnitude larger than the publicly available one for mitosis detection. Using this method, the total number of annotated mitotic figures in H\&E was solely limited by the number of restained slides available, not the amount of manual annotations. This is a very desirable property in the Computational Pathology field where manual annotations require plenty of resources and human expertise. Our work serves as a proof of concept to show that the combination of restaining, image analysis and registration can be used to automatically generate ground truth at scale when immunohistochemistry is the reference standard.

Staining variation between centers has long prohibited good generalization of algorithms to unseen data. In this work we applied a stain augmentation strategy that modifies the hematoxylin and eosin color channels directly, resulting in training samples with diverse and realistic H\&E stain variations. Our experimental results indicate that the use of H\&E-specific data augmentation and an ensemble of networks were key ingredients to drastically reduce the CNN's generalization error to unseen stain variations, i.e. transferring the performance obtained in WSIs from a single center to a cohort of WSIs from multiple centers. Furthermore, these results suggest that it is possible to train robust mitosis detectors without the need for assembling multicenter training cohorts or using stain standardization algorithms. More generally, we think that this combination of H\&E-specific data augmentation and ensembling could benefit other applications where inference is performed on H\&E WSIs, regardless of the tissue type.

High capacity CNNs typically exhibit top performance in a variety of tasks in the field of Computational Pathology, including mitosis detection. However, they come with a computational burden that can potentially compromise their applicability in daily practice. By using knowledge distillation, we massively reduced the computational requirements of the trained detector at inference time. In particular, we shrank the size of the distilled model 28 times (see Tab. \ref{tab:aug}), with a negligible performance loss. This reduction combined with the fully convolutional design of the distilled network enabled us to perform efficient dense prediction at gigapixel-scale, processing entire TUPAC WSIs at \SI{0.25}{\um/pixel} resolution in 30-45~\SI{}{\min}. 

On the task of individual mitosis detection in the \textit{TUPAC-aux-test} set, we obtained different precision scores from the TUPAC organizers (0.467) and our observer (0.707). We attribute this disagreement to two factors: (1) the method used to annotate the images, and (2) the small number of samples in the test set. According to TUPAC organizers, two pathologists independently traversed the image tiles and identified mitotic figures. Coincident detections were marked as reference standard, and discording items were reviewed by a third pathologist. Notice that mitotic figures missed by both pathologists were never reported, potentially resulting in true mitotic figures not being annotated. This lack of recall could explain the high number of false positives initially detected by our network and later found to be true mitotic figures by an expert observer. Due to the small number of samples in the \textit{TUPAC-aux-test} set (34 tiles), this effect can cause a large distortion in the F1-score. For reproducibility considerations, we have included all detections in the Appendix (Fig. \ref{fig:telle8.pdf}). These results illustrate the difficulty of annotating mitotic figures manually, specifically in terms of recalling them throughout large tissue regions, and supports the idea of using PHH3 stained slides as an objective reference standard for the task of mitosis detection.

As a limitation of our method, we acknowledge the existence of some noise in the reference standard generated by analyzing the PHH3 WSIs and attribute it to three components: (1) the limited sensitivity of the PHH3 antibody (some late-stage mitotic figures were not highlighted, thus not even considered as candidates); (2) the limited specificity of the PHH3 antibody (many of the candidates turned out to be artifacts); and (3) the limited performance of \textit{CNN2} (F1-score of 0.9). This noise restricted our ability to detect small performance changes during training, potentially resulting in suboptimal model and/or hyper-parameter choices. More carefully PHH3 restaining process and improved training protocols could palliate this effect in the future.

In terms of future work, mitotic density at whole-slide level could be exploited to find the location of the tumor hotspot, potentially resulting in significant speedups in daily practice. Additionally, the same metric could be used to study tumor heterogeneity, e.g. by analyzing the distribution of active tumor fronts within the sample. More generally, our work could be extended into other areas of Computational Pathology beyond mitosis detection in breast tissue by: (1) adopting the combination of slide restaining, automatic image analysis and registration to create large-scale training datasets where immunohistochemistry is the reference standard; and (2) validating the proposed stain augmentation strategy in other applications that analyze H\&E WSIs. 

\section*{Appendix}
\label{sec:appendix}

\subsection{Theory of color representation}

The Lambert-Beer law describes the relation between the amount of light absorbed by a specimen and the amount of stain present on it:

\begin{equation}\label{eq:lambertbeer}
\frac{I_i}{I_{0,i}} = \exp{(-A c_i)}\ ,
\end{equation}

where $I_i$ is the radiant flux emitted by the specimen, $I_{0,i}$ is the radiant flux received by the specimen, $A$ is the amount of stain, $c_i$ is the absorption factor, and subscript $i$ indicates one of the RGB color channels.\\

Based on this law, we cannot establish a linear relationship between the relative absorption detected by a RGB camera ($I_i / I_{0,i}$) and the amount of stain present in a specimen. However, we can instead define the optical density ($OD$) per channel as follows:

\begin{equation}\label{eq:odc}
OD_i = -\log{\frac{I_i}{I_{0,i}}} = -A c_i\ .
\end{equation}

Each $OD$ vector describes a given stain in the $OD$-converted RGB color space. For example, measurements of a specimen stained with hematoxylin resulted in $OD$ values of 0.18, 0.20 and 0.08 for each of the RGB channels, respectively \cite{ruifrok2001quantification}. \\

By measuring the relative absorption for each RGB channel on slides stained with a single stain, Ruifrok et al. \cite{ruifrok2001quantification} quantified these $OD$ vectors for hematoxylin, eosin and DAB (HED) stains. We can group these vectors into $M$, a 3 by 3 matrix representing a linear relationship between $OD$-converted pixels and the HED stain space. To achieve a correct balancing of the absorption factor for each stain, we divide each $OD$ vector in $M$ by its total length.\\

Therefore, a particular set of $OD$-converted pixels $y$ can be described as follows: 

\begin{equation}\label{eq:y}
y = x M\ ,
\end{equation}

where $x$ is a 1 by 3 vector representing the amount of stain (e.g. hematoxylin, eosin and DAB) per pixel, and $M$ is the normalized $OD$ matrix for the given combination of stains. Since we are interested in obtaining $x$, we only need to invert $M$:

\begin{equation}\label{eq:x}
x = y M^{-1}\ .
\end{equation}

\subsection{Color stain augmentation algorithm}

Given an RGB image patch $P \in \mathbb{R}^{MxMx3}$ reshaped into $P \in \mathbb{R}^{Nx3}$ with $N$ RGB pixels and the normalized $OD$ matrix $M \in \mathbb{R}^{3x3}$ for hematoxylin, eosin and DAB, we apply equations \ref{eq:odc} and \ref{eq:x} to transform the patch from RGB to HED color space as follows:

\begin{equation}\label{eq:patch}
S = -\log{(P + \epsilon)} M^{-1}\ ,
\end{equation}

where $S \in \mathbb{R}^{Nx3}$ is the transformed patch in HED color space and $\epsilon$ is a positive bias to avoid numerical errors. We simulate alternative stain intensities by stochastically modifying each stain component:

\begin{equation}\label{eq:patch_aug}
S'_i = \alpha_i S_i  + \beta_i\ ,
\end{equation}

where $S' \in \mathbb{R}^{Nx3}$ is the augmented patch in HED color space, subscript $i$ represents each stain channel, $\alpha_i$ is drawn from a uniform distribution $U(1-\sigma, 1+\sigma)$, $\beta_i$ is drawn from a uniform distribution $U(-\sigma, \sigma)$, and typically $\sigma = 0.05$.\\

To obtain an RGB representation of the augmented patch $S'$, we invert the operations described in equation \ref{eq:patch}:

\begin{equation}\label{eq:aug_rgb}
P' = \exp{(-S'M)} - \epsilon\ ,
\end{equation}

where $P' \in \mathbb{R}^{Nx3}$ is the augmented patch in RGB color space. Finally, we reshape $P'$ into $P' \in \mathbb{R}^{MxMx3}$ to match the original shape of the patch.

\begin{figure*}[h!]
\captionsetup{font=small}
\centering\includegraphics[width=1.0\linewidth]{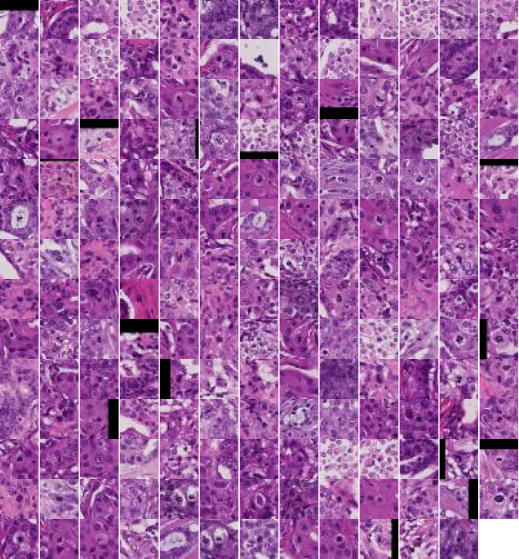}
\caption{
Mitosis detections in the \textit{TUPAC-aux-test} dataset identified by \textit{CNN5}. These patches were classified as containing a mitotic figure or not by a resident pathologist. The observer classified 128 out of 181 detections as true positives, resulting in a precision score of 0.707 for the automatic detector. 
}
\label{fig:telle8.pdf}
\end{figure*}

\ifCLASSOPTIONcaptionsoff
  \newpage
\fi

\section*{Acknowledgment}
\label{sec:acknowledgements}

This study was financed by a grant from the Radboud Institute of Health Sciences (RIHS), Nijmegen, The Netherlands. The authors would like to thank Mitko Veta, organizer of the TUPAC Challenge, for evaluating our predictions in the test set of the TUPAC dataset; NVIDIA Corporation for donating a Titan X GPU card for our research; and the developers of Theano \cite{2016arXiv160502688short} and Lasagne \cite{lasagne}, the open source tools that we used to run our deep learning experiments.

\bibliographystyle{IEEEtran}
\bibliography{IEEEexample}

\end{document}